\documentclass[final]{cvpr}

\makeatletter
\let\ESO@isMEMOIR\relax
\let\AtPageUpperLeft\relax

\let\AtTextUpperLeft\relax

\let\ESO@HookI\relax
\let\AddToShipoutPicture\relax

\let\ESO@HookII\relax
\let\ESO@HookIII\relax
\let\ESO@gridunit\relax
\let\ESO@labelfactor\relax
\let\ESO@gridunitname\relax
\let\ESO@griddelta\relax
\let\ESO@griddeltaY\relax
\let\ESO@gridDelta\relax
\let\ESO@gridDeltaY\relax
\let\ESO@gridcolor\relax
\let\ESO@subgridcolor\relax
\let\ESO@subgridstyle\relax
\let\ESO@gap\relax
\let\ESO@yoffsetI\relax
\let\ESO@yoffsetII\relax
\let\ESO@gridlines\relax
\let\ESO@subgridlines\relax
\let\ESO@hline\relax
\let\ESO@vline\relax
\let\ESO@Hline\relax
\let\ESO@Vline\relax
\let\ESO@fcolorbox\relax
\let\ESO@color\relax
\let\ESO@colorbox\relax
\let\gridSetup\relax
\let\ESO@div\relax
\let\ESO@gridpicture\relax
\makeatother
\usepackage{pdfpages}

\usepackage{times}
\usepackage{epsfig}
\usepackage{graphicx}
\usepackage{amsmath}
\usepackage{amssymb}

\usepackage[utf8]{inputenc} 
\usepackage[T1]{fontenc}    
\usepackage{url}            
\usepackage{booktabs}       
\usepackage{amsfonts}       
\usepackage{nicefrac}       
\usepackage{microtype}      
\usepackage{lipsum}		
\usepackage{graphicx}
\usepackage[numbers]{natbib}
\usepackage{doi}
\usepackage{bm}
\usepackage{multirow}
\usepackage{bbm}
\usepackage{amsmath} 

\renewcommand{\vec}[1]{\boldsymbol{#1}}

\newcommand{\given}{\, | \,}

\renewcommand{\to}{\longrightarrow}

\title{Monocular Depth Estimation via Listwise Ranking\\ using the Plackett-Luce Model}

\author{
Julian~Lienen$^{1,}$\thanks{Corresponding author: {\tt julian.lienen@upb.de}.} \quad
Eyke~H\"{u}llermeier$^2$ \quad
Ralph~Ewerth$^{3,4}$ \quad Nils~Nommensen$^{3,4}$ \\
$^1$ Paderborn University \quad  $^2$ University of Munich (LMU)
\\ $^3$ L3S Research Center, Leibniz University Hannover \quad $^4$ TIB Hannover \\
}

\def\cvprPaperID{10091}
\def\confYear{CVPR 2021}

\begin{document}

\maketitle

\begin{abstract}
    In many real-world applications, the relative depth of objects in an image is crucial for scene understanding. Recent approaches mainly tackle the problem of depth prediction in monocular images by treating the problem as a regression task. Yet, being interested in an \emph{order relation} in the first place, ranking methods suggest themselves as a natural alternative to regression, and indeed, ranking approaches leveraging pairwise comparisons as training information (``object A is closer to the camera than B'') have shown promising performance on this problem. 
    In this paper, we elaborate on the use of so-called \emph{listwise} ranking as a generalization of the pairwise approach. 
    Our method is based on the Plackett-Luce (PL) model, a probability distribution on rankings, which we combine with a state-of-the-art neural network architecture and a simple sampling strategy to reduce training complexity. Moreover, taking advantage of the representation of PL as a random utility model, the proposed predictor offers a natural way to recover (shift-invariant) metric depth information from ranking-only data provided at training time.
    An empirical evaluation on several benchmark datasets in a ``zero-shot'' setting demonstrates the effectiveness of our approach compared to existing ranking and regression methods.
\end{abstract}

\section{Introduction}

Estimating depth in monocular images constitutes a problem of practical importance when aiming to understand the geometry of a scene, e.g., in autonomous driving systems or for augmented reality applications. Due to its ill-posed nature, methods approaching this problem nowadays typically incorporate complex models, trained on large amounts of data using machine learning methods.

The majority of existing approaches tackles depth estimation (whether per-pixel or per-object) as a regression problem,  i.e., as the problem of learning a model to predict a (pseudo-)metric map (e.g., \cite{Alhashim2018,eigen2014depth, Laina2016DeeperDP, Lee2019FromBT}).
However, on the one hand, accurate prediction of metric depth actually depends on the intrinsic camera parameters, which are often not available. On the other hand, instead of predicting absolute depth, it is often enough to predict the \emph{relative} depth of pixels or higher level concepts (such as objects), that is, to sort them from closest to farthest away from the camera. 

One may then argue that regression is solving an unnecessarily difficult task, and rather advocate a formalization of depth estimation as a \emph{ranking} task \cite{pl11}. So-called ``learning-to-rank'' methods can be used to minimize suitable performance metrics based on relative errors. 
As absolute depth measurements are not necessarily needed, ranking has the additional advantage that it potentially allows for learning from weaker training information.
This includes depth annotations that are not metric but can be regarded as pseudo-metric data, e.g., disparity maps constructed from stereo images or videos \cite{Chen2019LearningSD,monodepth2, Li2018MegaDepthLS}, or human-annotated data \cite{NIPS2016_6489,DBLP:conf/cvpr/ChenQFKHD20}. 
Without the need for metric RGB-D data produced by depth sensors, the diversity of training datasets can be drastically increased due to cheaper data acquisition \cite{Xian2018MonocularRD}.

Existing ranking methods are essentially based on pairwise comparisons of the form ``object A is closer to the camera than B'' \cite{NIPS2016_6489, Xian2018MonocularRD, Xian2020StructureGuidedRL,Zoran2015LearningOR}. Pairwise relations of that kind are sampled from a depth map as training information, and predictive models are induced by minimizing pairwise ranking losses. While these approaches have proven effective, the quadratic number of possible pairs that can be constructed renders them rather inefficient and necessitates sophisticated sampling strategies to eliminate less informative pairs \cite{Xian2020StructureGuidedRL}. Besides, breaking a linear order into pairwise comparisons necessarily comes with a certain loss of information. In particular, information about the transitivity of order relations, which is implicitly contained in a linear order, will be lost.

To avoid these drawbacks, so-called ``listwise ranking'' \cite{xia2008listwise} has been proposed as an alternative to pairwise methods. In the listwise approach, higher order rankings of arbitrary length can be considered as training information. In this paper, we elaborate on the use of listwise ranking for depth estimation in images. More specifically, we propose a listwise ranking method based on the well-known Plackett-Luce (PL) model \cite{Luce59,Plackett1975TheAO}, which allows for learning probability distributions over rankings from pseudo-metric data. Moreover, taking advantage of the representation of PL as a random utility model \cite{DBLP:conf/nips/SoufianiPX12}, we suggest a natural way to recover translation-invariant approximations of the underlying metric depth information. Along with that, we propose a state-of-the-art neural network architecture as a backbone, together with a simple sampling strategy to construct training examples from raw pseudo-depth data. 

In a zero-shot evaluation, where we compare models on data not considered for training, we study the cross-dataset performance of our model and compare it with state-of-the-art approaches. Thereby, we demonstrate that listwise ranking is an effective approach for rank-based error minimization, and our model constitutes an appropriate choice for the prediction of depth orders in unseen scenes, as well as providing promising results in recovering metric depth.

\section{Related Work}

In learning to rank, the goal is to infer ranking models from training data in the form of rankings (permutations) of individual items. According to a rough categorization of methods, one can distinguish between pointwise, pairwise, and listwise approaches \cite{Liu2010LearningTR}. While single items are considered as training examples in pointwise learning-to-rank methods, relations between items are typically used as training examples in the other categories, either relations of order two (pairwise) or arbitrary length (listwise). In the case of pointwise learning-to-rank, examples are usually annotated by a score that determines their individual usefulness, from which, for instance, regression models can be induced. For pairwise approaches, where examples are typically given as single relations among two items, existing methods range from SVM-based classifiers \cite{rankingsvm} to boosting methods \cite{rankboost} and ranking networks \cite{RankNet}. Similarly, several listwise ranking methods have been proposed, in which examples are represented by higher order (potentially partial) item rankings. One of the most well-known representative is ListMLE \cite{xia2008listwise}, a maximum likelihood estimation method to infer Plackett-Luce probability distributions over rankings.

Several approaches to tackle the problem of estimating depth in images using relative depth information for training have been proposed. Among the first, Zoran et al.\ \cite{Zoran2015LearningOR} classify individual point pairs from an image, which are then combined into a global solution for a complete dense map over all image pixels. Following a similar motivation, Chen et al.\ \cite{NIPS2016_6489} train a deep neural network architecture by using a pairwise ranking loss, directly predicting a dense-map in an end-to-end fashion. This approach has also been adopted in subsequent works and improved in various directions, for example by using a different model architecture \cite{Xian2018MonocularRD}, additional data \cite{Chen2019LearningSD}, or improved sampling strategy \cite{Xian2020StructureGuidedRL}. Furthermore, Ewerth et al.\ \cite{Ewerth2017EstimatingRD} propose a method to estimate relative depth using a RankBoost model. Alternative approaches also exploit ordinal depth information \cite{Li2018MegaDepthLS}, either directly or to pretrain regression models \cite{cao2020monocular}. 

To learn models that work well for arbitrary scenes, e.g., in both indoor and outdoor scenarios, diversity of training data is crucial. Commonly used metric data produced by depth sensors typically provide limited diversity, e.g., NYUD-v2 \cite{NYUDV2} with indoor-only or KITTI \cite{Geiger2013IJRR} with only street scenes. Since maximal depth capacities of sensors constrain the recognizable depth, they fail to capture scenes ``in the wild''. This is why Chen et al.\ \cite{NIPS2016_6489} propose a human-annotated dataset with pairwise point samples, for which the ``closer-to-camera'' relation is captured. However, as it provides ground truth information for only two points in each image, and the human annotation process is quite costly, other strategies aiming to automatically extract depth information have been proposed. For instance, stereo images \cite{Xian2018MonocularRD} or sequences of images in videos \cite{Chen2019LearningSD} have been facilitated to predict structural disparity maps from the motion of elements. Combinations of such methods have been considered, too \cite{Wang2019WebSV}. As none of them delivers metric information per pixel, the information produced must be considered as pseudo-depth, which, as previously explained, is still sufficient for depth relations. Although scale-invariant regression methods are also capable of learning from such data \cite{Li2018MegaDepthLS,ranftl2020towards}, their ability to generalize to new datasets with structurally different scenes is fairly limited, at least for the task of depth ordering, as our empirical evaluation will confirm later on.

\section{Plackett-Luce Model for Depth Estimation}

\begin{figure}[!thbp]
    \centering
    \includegraphics[width=0.9\linewidth]{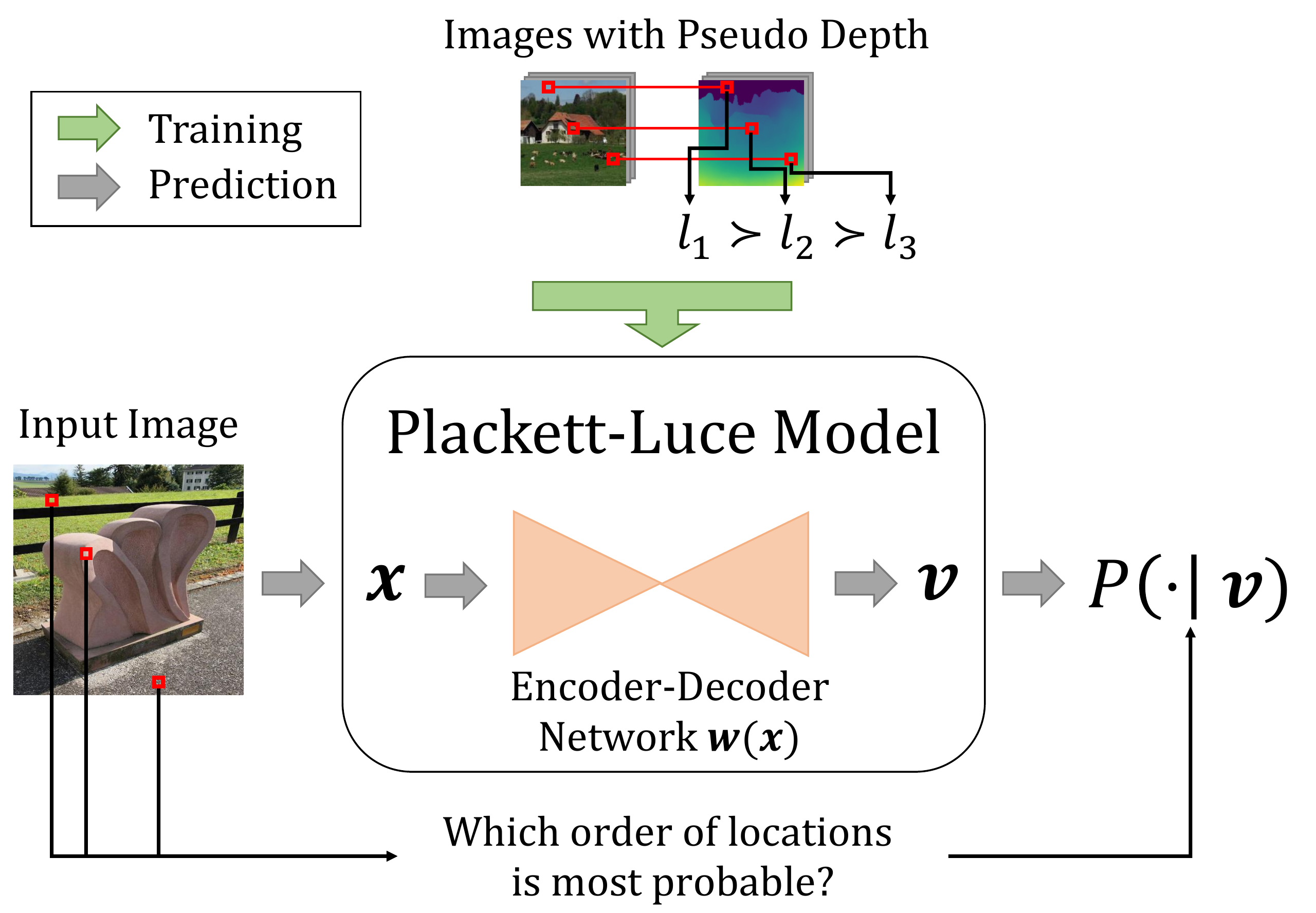}
   \caption{Overview of our method: The PL model incorporates a deep neural network to predict scores for each pixel in an input image, which are then turned into probabilities for rankings of queried image locations. For training, we sample rankings from images annotated by pseudo depth.}
\label{fig:plmodel}
\end{figure}

In the following, we introduce our proposal of a Plackett-Luce model for depth estimation as illustrated in Fig.\ \ref{fig:plmodel}, along with a description of the model architecture and sampling strategy to construct training examples from raw depth data.

\subsection{Problem Formulation}

We assume training information in the form of RGB images $I$ together with (pseudo-)depth annotations $D$, i.e., tuples $(I,D) \in \mathbb{R}^{h \times w \times 3} \times \mathbb{R}^{h \times w}$, where $h$ and $w$ denote the image height and width, respectively. Moreover, $D[l]$ denotes the (pseudo-)depth of a position $l \in \{1, \dots, h\} \times \{1, \dots, w\}$ identified by a height and width coordinate. Without loss of generality, lower values $D[l]$ encode shorter distances to the camera.

We are mainly interested in the order relation of the locations in an image  $I$ as induced by the (pseudo-)depth $D$. Formally, the relation between $n$ locations $M=\{l_1, l_2, \dots, l_n\}$ can be represented in terms of a permutation $\pi$ of $[n] := \{1, \ldots, n\}$ such that $D[l_{\pi(i)}] < D[l_{\pi(i+1)}]$ for $i \in \{1, \dots, n-1\}$. This permutation encodes the ranking 
$l_{\pi(1)} \succ l_{\pi(2)} \succ \dots \succ l_{\pi(n)}$, i.e., location $l_{\pi(1)}$ is closest, then $l_{\pi(2)}$, etc.
At query time, when $I$ is given but $D$ is not, the task of a rank-based depth estimation model is to predict the ``closer-to-camera'' relation $\succ$, that is, to produce an accurate order-preserving estimate of $D$. Formally, this estimate can again be represented in terms of a permutation, which is then compared to the ground truth permutation $\pi$.

\subsection{Listwise Depth Ranking}
\label{sec:method:listwise}

We model information about rankings in a \emph{probabilistic} way, which has several advantages, especially from a learning point of view (for example, it makes the problem amenable to general inference principles such as maximum likelihood estimation).
A well-known probability model on rankings is the Plackett-Luce (PL) model, which is parameterized by a vector $\vec{v} = (v_1, \dots, v_K) \in \mathbb{R}^K_+$, where $K$ is the number of items (length of the ranking). Referring to the interpretation of a ranking in terms of a preferential order, the value $v_i$ is also called the (latent) utility of the $i^{th}$ item\,---\,subsequently, we shall use the more neutral notion of PL score or parameter. The probability of a permutation $\pi$ of $[K]$ is then given by
\begin{equation}\label{eq:plm}
    P(\pi \, | \, \vec{v}) = \prod_{i=1}^{K-1} \frac{v_{\pi(i)}}{\sum_{k=i}^K v_{\pi(k)}} \, ,
\end{equation}
where $\pi(i)$ is the index of the item on the $i^{th}$ rank. One easily verifies that, the larger the score $v_i$, the higher the probability that the $i^{th}$ item will show up on a top rank. Moreover, the mode of the distribution, i.e., the ranking with the highest probability, is obtained by sorting the items in decreasing order of their scores.

The PL model has the appealing property that each marginal of a PL model is again a PL model (with the same parameters). More specifically, if $J = \{ j_1, \ldots , j_k \} \subseteq [K]$ is a subset of the $K$ items, then the corresponding marginal of (\ref{eq:plm}) is a PL model with parameters $v_{j_1}, \ldots , v_{j_k}$. This property greatly facilitates learning and inference from possibly incomplete rankings that do not comprise all $K$ items. In fact, learning to rank with the PL model essentially comes down to estimating the score vector $\vec{v} = (v_1, \dots, v_K)$. 

In the case of depth estimation, items correspond to the pixels of an image, and the task of the learner is to predict the scores of these pixels. To make this possible, we assume that the score of a pixel can be expressed as a function of its context on the image. Thus,   
a parameter $v_i$ is defined through a function $\phi_i : \mathcal{X} \to \mathbb{R}$ on an input space $\mathcal{X}$ \cite{cheng2010label}, where $\mathcal{X} = \mathbb{R}^{h \times w \times 3}$ corresponds to the space of all possible images of size $h \times w$. Assuming all images to have the same size, we set the overall number of alternatives $K$ to $h \times w$. 

In the domain of depth estimation, the most obvious way to represent the functions $\phi_1, \ldots , \phi_K$ is to model them as a (joint) deep convolutional neural network. Thus, each function $\phi_i$ is represented in terms of a set of network parameters $\vec{w}_i$, a subset of the parameters $\vec{w}$ of the entire (joint) network. In the experimental section, different state-of-the-art model architectures will be assessed for that purpose.

For an image $\vec{x} \in \mathcal{X}$, let $\vec{w}(\vec{x})$ denote the output of the neural network under parameterization $\vec{w}$ and 
\begin{equation}\label{eq:vw}
(v_1, \ldots , v_K) = (\phi_1(\vec{x}) , \ldots , \phi_K(\vec{x})) = \exp(\vec{w}(\vec{x}))
\end{equation}
the induced (non-negative) PL parameters. Thus, the entire PL model for the image $\vec{x}$ is eventually specified by the network parameters $\vec{w}$. Given a ranking $\pi$ of (a subset of) the pixels of $\vec{x}$ as training information, one can thus determine the probability $P(\pi \given \vec{x}, \vec{w})$ of that ranking under $\vec{w}$ according to (\ref{eq:plm}). More generally, given training information in the form of a collection of images with rankings, $\{ (\vec{x}_i , \pi_i ) \}_{i=1}^L$, learning an optimal model can be realized as maximum likelihood estimation \cite{xia2008listwise}:
\begin{equation}
    \vec{w}^* \in \operatorname*{arg\,min}_{\vec{w}}  - \sum_{i=1}^L \log P(\pi \given \vec{x}, \vec{w}) \, .
\end{equation}

\subsection{Metric Depth Estimation}
\label{sec:method:metric}

Going beyond the prediction of rankings, one may wonder whether there is any possibility to recover metric depth information from a learned PL model. At first sight, this would be surprising, because the model is only trained on qualitative information in the form of rankings, and predicts probabilities instead of metric depth. Yet, the PL model also comprises a quantitative part, namely the scores $v_i$, which, as will be explained in the following, are in direct correspondence with the underlying metric information.   

The PL model is a specific random utility model (RUM) \cite{Mcfadden1980EconometricMF}. In this class of models, it is assumed that the true order $z_1 < z_2 < \ldots < z_n$ of $n$ real numbers\,---\,think of them as the true depth values of the pixels in a image\,---\,is ``randomized'' through (independent) measurement noise: Each value $z_i$ is replaced by the measurement $X_i = z_i + \epsilon_i$, where $\epsilon_i$ is an error term, and what is observed as a ranking is the order of the measurements $X_1, \ldots , X_n$. In particular, the true order relation $z_i < z_j$ between two items is reversed if the corresponding error terms satisfy $\epsilon_i - \epsilon_j > z_j - z_i$, and the smaller the distance $| z_i - z_j|$, the more likely such a mistake is going to happen. Thus, the probability of a ranking error is indicative of the distance between $z_i$ and $z_j$.

The PL model is obtained for the special case where the error terms $\epsilon_i$ follow a Gumbel distribution with fixed shape parameter \cite{DBLP:conf/nips/SoufianiPX12}.
More specifically, the so-called Thurstone model with parameters $z_1, \ldots , z_n$ is equivalent to the PL model (\ref{eq:plm}) with parameters $v_i = \exp( z_i)$, $i=1 , \ldots , n$. In the context of depth estimation, the model can thus be interpreted as follows: The true depth of the $i^{th}$ image object (pixel) is given by $z_i$, but due to measurement noise, these distances are not observed precisely. Accepting the assumption of a Gumbel distribution\footnote{This distribution looks similar to the normal distribution. Even if not provably correct, it is certainly not implausible.}, a PL model fitted to the observed (noisy) rankings of image objects yields estimates $\hat{v}_i$ of $v_i = \exp( z_i)$. Thus, a natural estimate of the underlying metric depth is given by $\hat{z}_i = \log ( \hat{v}_i )$.

We note that, since the PL model (\ref{eq:plm}) is invariant toward multiplicative scaling (i.e., $P(\pi \given \vec{v}) \equiv P(\pi \given \lambda \vec{v})$ for $\lambda > 0$), the parameter $\vec{v}$ can only be determined up to a multiplicative factor. Correspondingly, the parameter $\vec{z}$ can only be determined up to an additive constant. This is indeed plausible: Assuming that the probability of reversing the order of two image objects only depends on their true distance $| z_i - z_j|$, this probability will not change by shifting the entire scene (i.e., moving the camera closer or farther away). In addition to this shift invariance, there is also a scaling effect, albeit of a more indirect nature. This effect is caused by fixing the shape parameter of the Thurstone model to 1. Therefore, instead of a simple log-transformation, we shall use an affine transformation of the form $\hat{z} =  s \log( \hat{v}) + t$, with $s,t \in \mathbb{R}$ fitted to the image at hand.

\subsection{Model}
\label{sec:method:model}

Regarding the underlying neural network, taking an image $\vec{x}$ as input and producing $\vec{w}(\vec{x})$ as used in (\ref{eq:vw}) as output, we suggest two variants of our listwise ranking approach. The first one, dubbed \textit{PLDepthResNet}, uses the same model architecture as suggested by Xian et al. \cite{Xian2018MonocularRD}. As a second model, by consideration of recent neural architecture research, we propose \textit{PLDepthEffNet} as a closely related architecture relying on EfficientNet \cite{Tan2019EfficientNetRM} as backbone. Without further notice, the variant EfficientNetB5 is used as encoder, while the decoder part is a stack of repeating convolutional, BatchNormalization, ReLU and bilinear upsampling layers until the original shape is recovered. Similar to the model in \cite{Xian2018MonocularRD}, different scale features from the encoder branch are fed into the corresponding levels of the decoder part. Instead of fusing these features by addition, we concatenate at the respective layers. As a result, we obtain a model with approximately $45$ million parameters for PLDepthEffNet, which is similar to the size of PLDepthResNet with $42$ million parameters, while increasing the model performance at the same time (cf. empirical evaluation).

For both PLDepthResNet and PLDepthEffNet, we use encoders pretrained on ImageNet. Consequently, we standardize input images to match the preprocessing on ImageNet. 
During training, we freeze the encoder part and only allow the BatchNormalization layers to adjust to the new input data as typically done in transfer learning.

\subsection{Sampling}
\label{sec:method:sampling}

In the past, different strategies to construct pairwise relations from raw depth data have been proposed, including superpixel sampling \cite{Zoran2015LearningOR}, random sampling \cite{NIPS2016_6489}, and combinations of multiple structure-guided strategies \cite{Xian2020StructureGuidedRL}. According to Xian et al. \cite{Xian2020StructureGuidedRL}, random sampling of pairwise relations from raw depth data may harm the model's performance, due to training on uninformative or even misleading examples. Even worse, due to imprecision in the ground truth data, the risk of incorrectly ordered items increases with larger samples.

To address these issues, we propose a random sampling strategy that is almost as simple as pure random sampling, and which allows for incorporating the depth structure of the given image while leading to a relatively low training complexity. For $R$ $n$-ary rankings to be queried per training tuple $(I, D)$, $N \cdot R$ item sets $M$ with $n$ individual image locations are sampled, where $N > 1$ is a parameter. For each ranking set $M$, we order all image locations $l$ by $D[l]$ to construct a ground truth permutation $\pi$. Given $\pi$, we sum up all pairwise depth differences 
$| D[l_{\pi(i)}] - D[l_{\pi(i + 1)}]|$, $i \in [n-1]$.
Afterwards, we sort all $N \cdot R$ rankings per image in a decreasing order according to this sum of depth difference and select the top $R$ rankings as training examples. This way, we consider those rankings that seem to be most informative, since their relative depth values are maximized among the samples. Other strategies, such as the minimum among all pairwise depth differences in a ranking, are of course also possible as a proxy of the amount of information.

It is worth to mention that the Plackett-Luce model does not support partial rankings, i.e., neither allows for ties nor incomparability between items. Thus, as opposed to strategies incorporating equality relations, as e.g.\ \cite{NIPS2016_6489}, such relations are not explicitly considered here.
To avoid sampling point pairs that are almost equally far away from the camera, we add a penalty of $-10$ to the depth difference sum for each compared image location pair $l_1$ and $l_2$ if their depth difference is such that
$\max\left\{ \frac{D[l_1]}{D[l_2]}, \frac{D[l_2]}{D[l_1]} \right\} < 1 + \tau$,
where the parameter $\tau$ is set to $\tau = 0.03$ in our experiments.

\section{Experiments}

To demonstrate the effectiveness of our method, we conduct an exhaustive empirical evaluation on several benchmark datasets. Before presenting the results, we first introduce the datasets, followed by a brief description of the baseline methods and metrics used for assessment.

\subsection{Datasets}

To train our models, we use the recently introduced pseudo-metric ``High-Resolution Web Stereo Image'' ($\textit{HR-WSI}$) dataset \cite{Xian2020StructureGuidedRL}. It consists of $20,378$ diverse, high resolution images annotated with pseudo-depth maps generated from flow predictions. For hyperparameter optimization, a separate set of $400$ images was used. Since the flow predictions provided as depth annotation failed for some image regions, a consistency mask is attached to each prediction to allow for sampling only from pixels that provide a reasonable depth value. To this end, a forward-backward flow consistency check has been applied. Furthermore, the annotations have been preprocessed to also assign a constant depth value to sky regions. Despite its relatively small size, we found this dataset to provide highly informative image and depth pairs to learn from.

In the experiments, we compare our model to various baselines in a ``zero-shot'' generalization study on datasets that were not used within the training processes. Thus, we follow the basic evaluation scheme by Ranftl et al. \cite{ranftl2020towards}. As datasets, we consider \textit{Ibims} \cite{Koch2018EvaluationOC}, \textit{Sintel} \cite{Butler2012ANO}, \textit{DIODE} \cite{Vasiljevic2019DIODEAD}, and \textit{TUM} \cite{Sturm2012ABF}. In the supplementary material, we detail the characteristics of each dataset, such as their data diversity.
With this choice of benchmark targets, we capture indoor, outdoor, and computer generated scenes, which provides a good basis for assessing the generalization performance of different models, and their ability to predict depth orders in a wide variety of applications.

\subsection{Baselines}

We compare our PL-based approach to state-of-the-art depth estimation models using depth relations as training information. To this end, we consider the ResNet-based model trained on ``Relative Depth from Web'' (ReDWeb), ``Depth in the Wild'' (DIW), and YouTube3D as described by Chen et al. \cite{Chen2019LearningSD}, hereinafter referred to as YouTube3D, and the same model as used by Xian et al. \cite{Xian2020StructureGuidedRL} trained on HR-WSI (referred to as Xian 2020). Both approaches have shown compelling generalization performance, corroborating our motivation to use relative data for supervision.

Besides models trained on relative depth information, regression models are obviously also capable of inferring rankings, simply by sorting the image locations based on their values in a predicted dense depth map. Therefore, we consider state-of-the-art (pseudo-)regression methods as additional baselines, namely, DenseDepth~\cite{Alhashim2018}, BTS~\cite{Lee2019FromBT}, MegaDepth~\cite{Li2018MegaDepthLS}, MannequinChallenge (MC)~\cite{li2019learning}, and MiDaS~\cite{ranftl2020towards}. Furthermore, we also evaluated MonoDepth2 \cite{monodepth2} as a completely unsupervised resp.\ self-supervised method.

While we considered most baselines as described in the related work, let us note that the authors of MiDaS provide a model trained on approximately $2$ million examples, which is far more than most of the other methods we compare with. To account for this, we re-implemented their approach and retrained the model on HR-WSI for a fairer comparison. For a complete overview of all baselines, including a categorization of the respective training data diversity, we refer to the supplementary material.

\subsection{Metrics}

To evaluate our models, we report the ``ordinal error'' on sampled point pairs as done by Xian et al. \cite{Xian2020StructureGuidedRL}. For two points $l_1$ and $l_2$ sampled from an example $(I, D)$, with $I$ being the image and $D$ a dense (pseudo-)depth map as specified before, the ground truth ordinal relation $r(l_1, l_2, D)$ is given by $+1$ for $D[l_1] > D[l_2]$, $-1$ for $D[l_2] > D[l_1]$ and $0$ otherwise.
The ordinal error is then given by 
\begin{equation}
    ord(\mathcal{D}) = \frac{1}{|\mathcal{D}|} \sum_{(I, D, l_1, l_2) \in \mathcal{D}} \!\!\!\!\!\!\!\mathbbm{1}\big( r(l_1, l_2, D) \neq r(l_1, l_2, f(I)) \big),
\end{equation}
where $f$ is the function predicting depth or, in the case of a PL model, scores for each pixel of the input image $I$, resulting in a dense depth map just as given by $D$, and $\mathcal{D}$ denotes the set of all point pairs sampled from the test dataset images and depth maps.

As already noted, we omit all equal pairs, i.e., relations with $r(\cdot, \cdot, \cdot)=0$. Hence, we report $ord$ on unequal pairs only without any equality thresholding. 
Thus, there is no need to rely on re-scaling and -translating as done in \cite{ranftl2020towards} and \cite{Xian2020StructureGuidedRL} to identify reasonable equality thresholds, which comes with additional complications for the evaluation process.

Often, depth orders have varying priorities, i.e., closer elements are more critical for correct ordering than elements far away from the camera. For example, an autonomous vehicle has less time to react to elements very close to the car and must rely on valid input for safe interactions. This is reflected by metrics like the discounted cumulative gain (DCG), which measures the usefulness of rankings by accumulating graded relevances of ranking items discounted with decreasing rank. More precisely, for every image location $l$ associated with a dense depth map $D$, we set the relevance score of $l$ in $D$ to $rel(l, D) = \frac{1}{D[l] + 1}$. Given these scores, we can specify the DCG score for a ranking $l_{\pi(1)} \succ l_{\pi(2)} \succ \dots \succ l_{\pi(n)}$ by
\begin{equation}\label{eq:dcg}
    DCG(\pi, D) = \sum_{i = 1}^{n} \frac{rel(l_{\pi(i)}, D)}{\log_2 (i +1)} \, .
\end{equation}
For our experiments, we used the normalized DCG (nDCG), which divides (\ref{eq:dcg}) by the best DCG 
possible on $D$.

For the metric comparison, we assess the root-mean-square error (RMSE) between the dense ground truth and predicted depth maps and the percentage of predictions $\hat{z}$  such that $\max\left(\frac{\hat{z}}{z}, \frac{z}{\hat{z}}\right) = \delta > 1.25$ for the ground truth depth $z$. To calculate the metrics, we normalized the given ground truth scores by the maximum depth capacity of the corresponding dataset (cf. the dataset characteristics in the supplement) to obtain error values on a similar scale.

\subsection{Results}

To show the effectiveness of our method proposal,
we first compare different losses using the same model architecture and training dataset, followed by a comparison of our method to the  baselines. 
Every reported result is the average of three runs with different randomization seeds.

\subsubsection{Loss Comparison}

There are many experimental studies in the literature showing improved performance of a method, but not isolating the key factors contributing to the improvement, e.g., the neural network architecture, loss function, training procedure, training data, etc. To assess the influence of a listwise approach to ranking more clearly, we evaluate three methods trained on the same data and with the same neural network architecture, namely (scale-invariant, SI) regression, pairwise, and listwise ranking. It is true that the model, loss, and data may strongly interact with each other (i.e., a loss might work well with a certain architecture on a particular dataset, while the same architecture may harm the performance of a different method). Nevertheless, we found that the ResNet-based architecture as proposed by Xian et al.~\cite{Xian2018MonocularRD} and subsequently also used in~\cite{ranftl2020towards} serves as a good basis for a fair comparison. 

For our experiments, we re-implemented the SI mean-squared error loss as also used in MiDaS and the pairwise ranking loss as described in~\cite{NIPS2016_6489} and~\cite{Xian2018MonocularRD}. As training information, we used HR-WSI as a state-of-the-art diverse pseudo-depth dataset. We refer to the supplement for a detailed description of all hyperparameters.

All three methods require different sampling strategies: While the SI-regression uses the complete (masked) image, pair- and listwise methods involve different amounts of sampled points selected per ranking. For a fair comparison, we adopted the number of sampled rankings in the listwise case to the number of drawn pairwise relations, such that one approach does not see much more points than the other during training. In the case of pairwise rankings, we randomly sampled $1$k point pairs per image and epoch, resulting in a maximum of $2$k seen points per image and epoch. For our listwise approach, we found a size of $5$ to achieve a good trade-off between highly informative rankings and efficient training. Hence, we sampled $400$ rankings of ranking size $5$ per image and epoch. Here, we explicitly stick to random-only sampling to alleviate side effects.

Table \ref{tab:experiments:results:losses} presents the results of the method comparison on $50$k randomly sampled location pairs per image. As can be seen, the relative models outperform the SI-regression method, suggesting to serve as a better surrogate loss for optimizing the ordinal error. Moreover, our listwise approach seems to perform slightly better than the pairwise approach, although the difference does not appear to be significant.

\begin{table}[t]
    \centering
        \caption{Ordinal errors on $50$k randomly sampled pairs per loss, using the architecture from \cite{Xian2020StructureGuidedRL} trained on HR-WSI (lower is better).}
    \label{tab:experiments:results:losses}
    \resizebox{0.93\columnwidth}{!}{
    \begin{tabular}{l|cccc|c}
    \toprule
         Loss & Ibims & Sintel & DIODE & TUM & Avg. Rank  \\
         \midrule
         SI-Regression & 0.308 & 0.311 & 0.334 & 0.222 & 3 \\
         Pairwise & 0.281 & 0.299 & 0.291 & \textbf{0.192} & 1.75 \\
         Listwise & \textbf{0.273} & \textbf{0.289} & \textbf{0.285} & 0.218 & \textbf{1.25} \\
         \bottomrule
    \end{tabular}}
\end{table}

\subsubsection{Ordinal Prediction}\label{sec:ordinal}

After having compared the loss function on a shared model and data level, we now analyze individual depth estimation models with regard to their ordinal error and nDCG performance as trained by the respective authors,
who made an attempt at optimizing the interplay between data, network architectures, and training procedures.

For the baseline models, we used the best provided pretrained models by the authors or, if official implementations were not available, by popular and carefully tested re-implementations. For our PL models, we kept most of the training hyperparameters the same (see supplementary for more details). Within our sampling strategy, we set the factor $N=5$ (cf.\ Section \ref{sec:method:sampling}). For MiDaS, we also used our proposed EfficientNet-based architecture, which delivers superior performance compared to the formerly used architecture, for reasons of fairness. Here, as opposed to the version of MiDaS within the loss comparison, where we primary focused on comparing different problem considerations, we employ the trimmed absolute deviation loss providing the best performance among the regarded alternatives (cf.~\cite{ranftl2020towards}).

Table \ref{tab:experiments:results:ord} reports the individual ordinal errors on unequal relations for the four benchmark datasets, again on $50$k randomly sampled location pairs per image. As can be seen, our PLDepthEffNet achieves the lowest averaged rank over all datasets, while outperforming the other methods on half of the datasets at the same time, demonstrating the effectiveness of the listwise ranking approach to optimize the ordinal error metric. Supporting the observations made in the previous experiment, the generalization capabilities of MegaDepth as another scale-invariant regression method, even by having access to over $600$k diverse instances, to correctly rank elements are fairly limited. Moreover, in agreement with the previous results, the ranking approaches are consistently among the best models, suggesting ranking losses to be the favorite choice as surrogates for ordinal error minimization.

\begin{table}[t]
	\centering
		\caption{Ordinal errors on benchmark datasets with $50$k randomly sampled relations for each image (lower is better).}
	\label{tab:experiments:results:ord}
	\resizebox{0.93\columnwidth}{!}{
	\begin{tabular}{l|cccc|c}
		\toprule
		Model & Ibims & Sintel & DIODE & TUM & Avg. Rank \\
		\midrule
		DenseDepth &  0.208 & 0.384 & 0.317 & 0.224 & 5.75 \\
		MegaDepth &  0.297 & 0.324 & 0.316 & 0.227 & 7.5 \\
		BTS & \textbf{0.190} & 0.384 & 0.323 & 0.251 & 6.25 \\
		MC &  0.272 & 0.387 & 0.378 & 0.206 & 7.25 \\
		MiDaS & 0.269 & 0.278 & 0.263 & 0.207 & 3.75 \\
		\midrule
		MonoDepth2 & 0.375 & 0.425 & 0.407 & 0.336 & 9.75 \\
		\midrule
		YouTube3D & 0.272 & 0.292 & 0.288 & 0.199 & 4.75 \\
		Xian 2020 & 0.225 & 0.278 & 0.263 & \textbf{0.184} & 2.25 \\
		\midrule
		PLDepthResNet & 0.245 & 0.284 & 0.277 & 0.213 & 4.75 \\
		PLDepthEffNet & 0.213 & \textbf{0.272} & \textbf{0.256} & 0.204 & \textbf{2} \\ 
		\bottomrule
	\end{tabular}}
\end{table}

Additionally, Table \ref{tab:experiments:results:ndcg} reports the results for nDCG as performance metric on $100$ randomly sampled rankings of size $500$ per image. In accordance with the ordinal errors, ranking methods are well suited to optimize this metric. Here, the top-$3$ models are all of that kind, with PLDepthEffNet slightly better performing than Xian 2020. 

\begin{table}[t]
	\centering
		\caption{nDCG on benchmark datasets with $100$ randomly sampled rankings of size $500$ for each image (higher is better).}
	\label{tab:experiments:results:ndcg}
	\resizebox{0.93\columnwidth}{!}{
	\begin{tabular}{l|cccc|c}
		\toprule
		Model & Ibims & Sintel & DIODE & TUM & Avg. Rank \\
		\midrule
		DenseDepth & 0.916 & 0.986 & 0.821 & 0.986 & 4.75 \\
		MegaDepth & 0.911 & 0.989 & 0.815 & 0.983 & 7.5 \\
		BTS & \textbf{0.918} & 0.986 & 0.825 & 0.983 & 4.75 \\
		MC & 0.908 & 0.986 & 0.828 & 0.987 & 5.5 \\
		MiDaS & 0.913 & 0.991 & 0.806 & 0.987 & 6.25 \\
		\midrule
		MonoDepth2 & 0.896 & 0.981 & \textbf{0.836} & 0.961 & 7.75 \\
		\midrule
		YouTube3D & 0.911 & 0.993 & 0.816 & 0.988 & 4.75 \\
		Xian 2020 & 0.916 & 0.993 & 0.817 & \textbf{0.990} & 2.75 \\
		\midrule
		PLDepthResNet & 0.914 & 0.993 & 0.817 & 0.985 & 5 \\
		PLDepthEffNet & 0.916 & \textbf{0.994} & 0.819 & 0.988 & \textbf{2.5} \\ 
		\bottomrule
	\end{tabular}}
\end{table}	

\subsubsection{Metric Prediction}

\begin{table*}[!t]
	\centering
		\caption{Evaluation results on benchmark datasets with regard to metric depth error measures (lower is better in both cases).}
	\label{tab:experiments:results:metric}
	\resizebox{0.805\textwidth}{!}{
	\begin{tabular}{l|cc|cc|cc|cc|cc}
	\toprule
		\multirow{2}{*}{Model} & \multicolumn{2}{c|}{Ibims} & \multicolumn{2}{c|}{Sintel} & \multicolumn{2}{c|}{DIODE} & \multicolumn{2}{c|}{TUM} & \multicolumn{2}{c}{Avg. Rank} \\
		& RMSE & $\delta > 1.25$ & RMSE & $\delta > 1.25$ & RMSE & $\delta > 1.25$ & RMSE & $\delta > 1.25$ & RMSE & $\delta > 1.25$ \\
		\midrule
		DenseDepth & \textbf{0.016} & 20.9 & 0.128 & 39.6 & 0.110 & 53.5 & 0.084 & 69.7 & 5.25 & 4.5 \\
		MegaDepth & 0.020 & 35.9 & 0.119 & 35.5 & 0.094 & 55.3 & 0.082 & 70.8 & 6 & 7 \\
		BTS & \textbf{0.016} & \textbf{18.9} & 0.133 & 41.8 & 0.112 & 54.4 & 0.089 & 72.4 & 7 & 6.25 \\
		MC & 0.018 & 31.3 & 0.128 & 38.8 & 0.120 & 58.7 & \textbf{0.074} & \textbf{67.8} & 5.25 & 5.5 \\
		MiDaS & 0.019 & 33.2 & \textbf{0.091} & \textbf{27.7} & \textbf{0.081} & 53.5 & 0.085 & 71.1 & 4 & 4.75 \\
		\midrule
		MonoDepth2 & 0.023 & 42.6 & 0.143 & 43.8 & 0.122 & 61.1 & 0.088 & 72.5 & 9.75 & 10 \\
		\midrule
		YouTube3D & 0.019 & 31.8 & 0.101 & 31.1 & 0.096 & 54.5 & 0.077 & 68.4 & 4.75 & 5.25 \\
		Xian 2020 & 0.018 & 31.5 & 0.096 & 30.5 & 0.085 & \textbf{51.4} & 0.080 & 69.4 & \textbf{3} & \textbf{3.25} \\
		\midrule 
		PLDepthResNet & 0.019 & 30.9  & 0.099 & 30.7 & 0.092 & 53.1 & 0.084 & 71.9 & 5 & 4.75 \\
		PLDepthEffNet & 0.017 & 29.1 & 0.093 & 29.3 & 0.085 & 52.7 & 0.083 & 71.6 & \textbf{3} & 3.5 \\
		\bottomrule
	\end{tabular}}
\end{table*}

As motivated theoretically in Section~\ref{sec:method:metric}, our method provides an interface to recover metric depth information approximated from observed rankings. Here, we compare our model to the baselines with regard to the two metric error measures RMSE and $\delta > 1.25$ using the same models as in Section \ref{sec:ordinal}. As all benchmark datasets have different scales and might be shifted arbitrarily, we rescale and shift the predictions to the resolution of the ground truth as described in \cite{ranftl2020towards} by optimizing a least-squares criterion.

The results are given in Table \ref{tab:experiments:results:metric}. As can be seen, although our model was solely trained on rankings, it is capable of recovering the underlying depth structure relatively precisely. Noteworthy, it is superior to all regression baselines and on a par with Xian 2020 for RMSE, although this ranking baseline additionally incorporates a smooth gradient loss term for sharp boundaries, directly accessing the metric depth information at training time. While it delivers the highest $\delta > 1.25$ accuracy, our approach still proves to be very competitive in this regard.

\begin{figure}[!t]
    \centering
    \includegraphics[width=\linewidth]{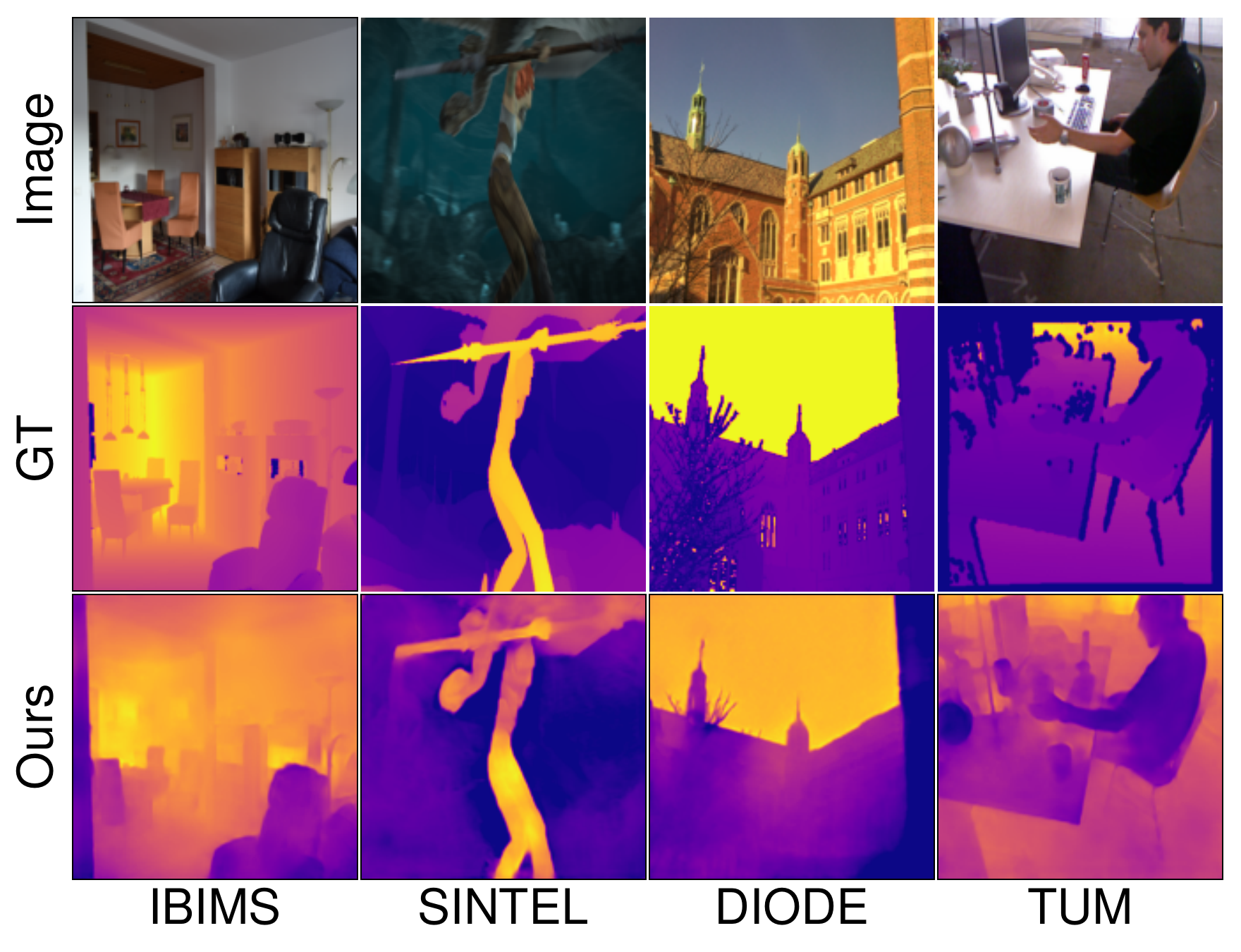}
   \caption{Sample predictions given by the reconstructed metric scores of the PLDepthEffNet model as used in the experiments.}
\label{fig:preds}
\end{figure}

Fig.\ \ref{fig:preds} shows exemplary predictions of our model. Obviously, the model is able to capture tiniest object details, such as tree branches in the image from DIODE, and predicting sharp object boundaries. This shows that, even with simple sampling strategies, listwise ranking is able to reflect and predict such small details, without any need for very complex strategies based on the depth structure of an image. 

\section{Conclusion}

We have proposed to tackle the problem of depth ordering in images as a listwise ranking problem, for which we employed a Plackett-Luce model tailored to the domain of monocular depth estimation. Thus, compared to estimating the exact depth values, we  solve an arguably simpler problem, at least if the goal is to minimize an ordinal error metric. Besides, compared to precise numerical data required by regression models for training, a ranking approach allows for leveraging weaker and more diverse training data. Although not directly trained on metric data, our model is capable of providing precise (shift-invariant) depth predictions, essentially by exploiting the relationship between the (latent) distance between image objects and the probability of reversing their order in a ranking.

Through an exhaustive zero-shot cross-dataset evaluation, we showed that our approach, combined with a state-of-the-art neural network as backbone, achieves superior ranking performance compared to previous approaches. In particular, it improves upon existing pairwise ranking methods, in spite of using a much simpler and more efficient sampling technique. Remarkably, our model also performs very competitive on metric error measures.

Motivated by these promising results, we plan to elaborate on further improvements of the listwise ranking approach. This includes an investigation of the effect of varying the ranking size, as well as an extension toward learning from partial rankings and incorporating equality relations. In addition, as we only applied random sampling so far, we plan to develop more sophisticated sampling strategies leading to more informative rankings to learn from. 

\noindent \textbf{Acknowledgement.} This work was supported by the German Research Foundation (DFG) under Grant\ 3050231323. Moreover, computational resources were provided by the Paderborn Center for Parallel Computing (PC$^2$).


{\small
\bibliographystyle{ieee_fullname}
\bibliography{bib_refs}
}

\clearpage



\section*{Supplementary material}

# – Supplementary Material –
# Monocular Depth Estimation via Listwise Ranking
# using the Plackett-Luce Model

## 1. Benchmark Dataset Characteristics

Table 1 shows the characteristics of the benchmark datasets, including their respective maximum depth values (capacity). We use these values to limit the recognized depth in the images as described in [9]. These values were also used to normalize the ground truth depths to get metric errors on a similar scale for all datasets.

Table 1. Properties of datasets being used within the zero-shot cross-dataset evaluation.

| Dataset | # Test Instances | Sensor Type | Diversity | Depth Capacity (in m) |
|---|---|---|---|---|
| Ibims [5] | 100 | Laser | Indoor | 50 |
| Sintel [2] | 1064 | Synthetic | Animated | 72 |
| DIODE [12] | 771 | Laser | Indoor, outdoor | 350 |
| TUM [10] | 1815 | Motion Parallax | Indoor | 10 |

Since the depth annotations of Sintel [2] are given in the inverse depth space, we did not apply the transformation $\frac{1}{D[l]+1}$ to calculate the nDCG scores as described in our paper (cf. Section 4.3). Instead, we directly used the already inverted depth values. Furthermore, in the case of TUM, we used the preprocessed dataset version as provided by the authors of [7]. Thereby, we considered the motion parallax (so-called "Plane-Plus-Parallax") depth map, which was constructed based on flow predictions between multiple views. We found these depth signals to provide a more precise and reliable source for calculating our error metrics compared to the originally provided Kinect sensor values.

## 2. Baseline Characteristics

Table 2 gives an overview of all baseline models considered within our empirical evaluation, together with the data being used for training. The diversity of the training data categorizes the individual datasets in terms of the variety of their captured scenes. MC represents a special case due to only incorporating images showing humans, but also indoor and outdoor.

Table 2. Baselines together with their training data considered within our empirical study.

| Model | Loss Class | Training set | # Examples | Training Data Diversity |
|---|---|---|---|---|
| DenseDepth [1] | (Scale-invariant) Regression | NYU | 50k | Low |
| MegaDepth [8] | | MegaDepth | 626k | High |
| BTS [6] | | NYU | 24k | Low |
| MC [7] | | MC | 136k | Medium |
| MiDaS [9] | | HR-WSI | 20k | High |
| MonoDepth2 [4] | Self-Sup. | KITTI | 40k | Low |
| YouTube3D [3] | Relative | RW+DIW+YT3D | 1219k | High |
| Xian 2020 [13] | | HR-WSI | 20k | High |

## 3. Experimental Details

For the loss comparison (cf. Section 4.4.1), we compared our model on the ResNet-based architecture (PLDepthResNet) to the scale-invariant regression [9] and pairwise ranking [13] approach. Thereby, we optimized all models for 50 epochs with Adam and a batch size of 40 on four Nvidia Titan RTX. For the scale-invariant regression and our PL model, we used an initial learning rate of 0.001 multiplied by $\sqrt{0.1}$ after 25 epochs, while the pairwise ranking approach used an initial learning rate of 0.01 with the same learning rate schedule. In all three cases, input images were resized to $448 \times 448$ and data has been augmented by horizontally flipping with a 50% chance.

In the second experimental study, where we trained our model on the proposed EfficientNet-based architecture (PLDepth-EffNet, cf. Fig. 3), we used a smaller initial learning rate of $0.0001$ with the other parameters kepth the same. For each image, we sampled $100$ rankings of size $5$ per epoch. We further evaluated the scale-invariant regression variant on the same model architecture, where we used the same hyperparameters as for PLDepthEffNet, but with a higher learning rate of $0.001$. The learning rate schedule was kept the same as for the previous experiments.

## 4. Additional Experiments on HR-WSI

As an additional experiment, we report the ordinal error, nDCG, RMSE and $\delta > 1.25$ as specified in the paper on the dataset HR-WSI [13]. Here, we compare only models being trained on this dataset, namely our PL model, MiDaS and Xian 2020. The presented results were computed on the separate validation set of $400$ instances, which was used for model optimization of the mentioned approaches. Thereby, we consider the same trained models as used for the ordinal and metric error calculation in Section 4.4.2 and 4.4.3 of our paper.

Table 3 shows the averaged results for three runs. As can be seen, our model is superior with regard to most of the metrics. Only for the nDCG, the scale-invariant regression variant MiDaS turns out to be superior, although only slightly. Fig. 1 further shows exemplary predictions on HR-WSI.

Table 3. Results on the validation split of HR-WSI for the models being trained on the corresponding training split. The same experimental settings as for the model comparison apply here. ↓ refers to "lower is better", while ↑ denotes the opposite.

| Model | Ord. Err. (↓) | nDCG (↑) | RMSE (↓) | $\delta > 1.25$ (↓) |
|---|---|---|---|---|
| MiDaS [9] | 0.192 | **0.839** | 0.088 | 0.294 |
| Xian 2020 [13] | 0.166 | 0.838 | 0.154 | 0.558 |
| PLDepthEffNet | **0.164** | 0.837 | **0.069** | **0.192** |

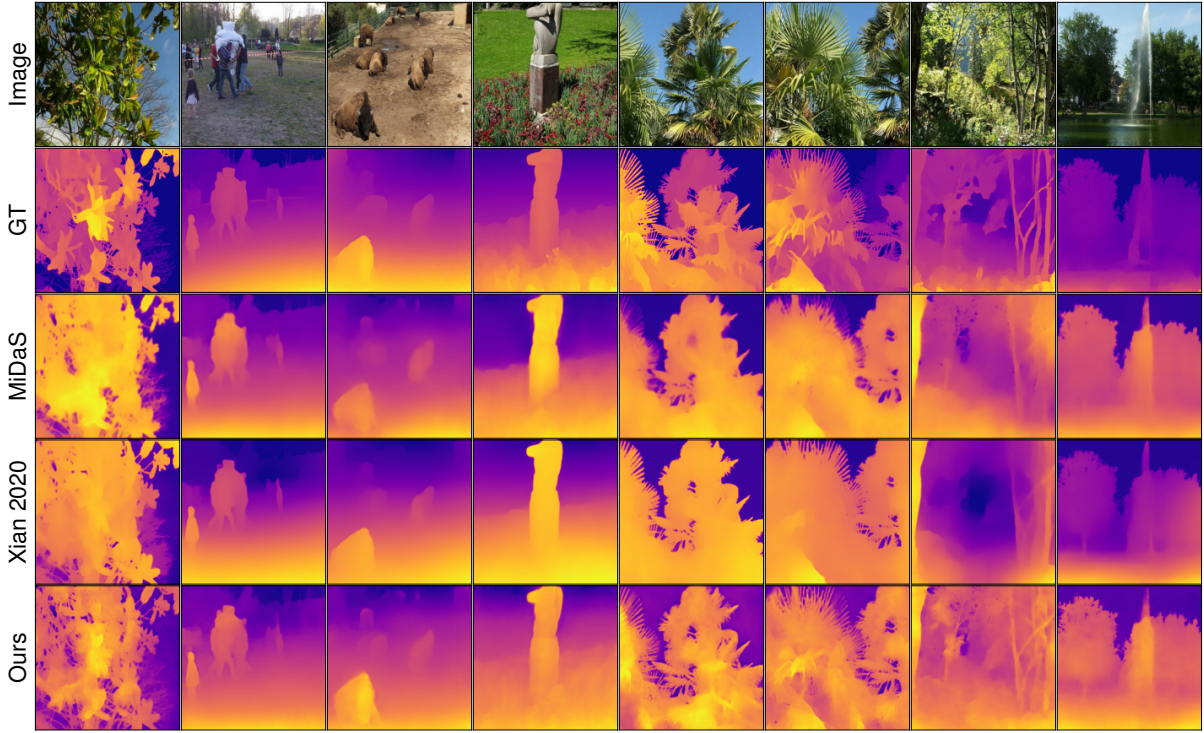

Figure 1. Exemplary predictions of the models trained on HR-WSI for validation set samples.

# 5. Additional Model Predictions

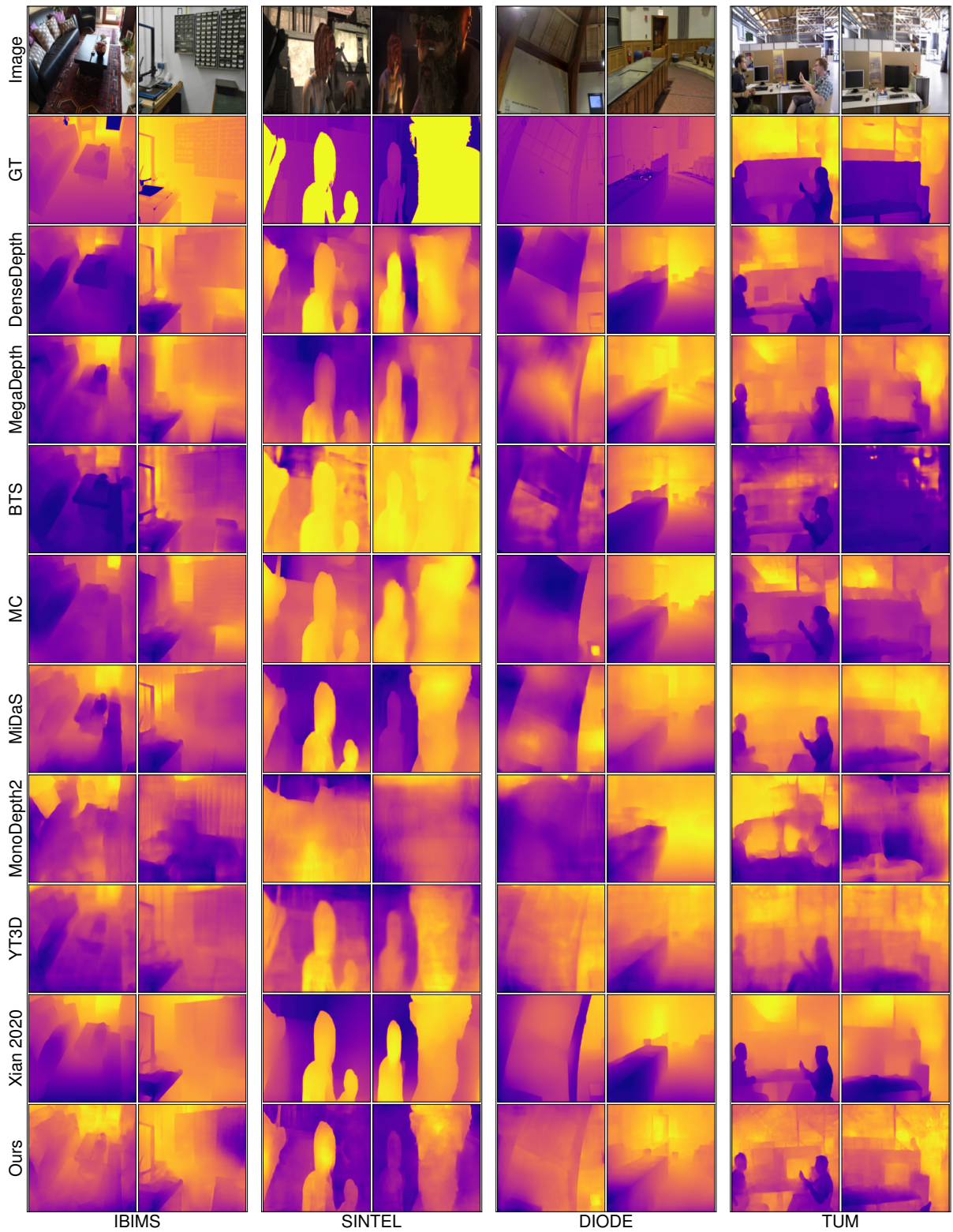

Figure 2. Model predictions for samples of the benchmark datasets (rescaled and shifted acc. to the ground truth depth values).

## 6. PLDepthEffNet Model Architecture

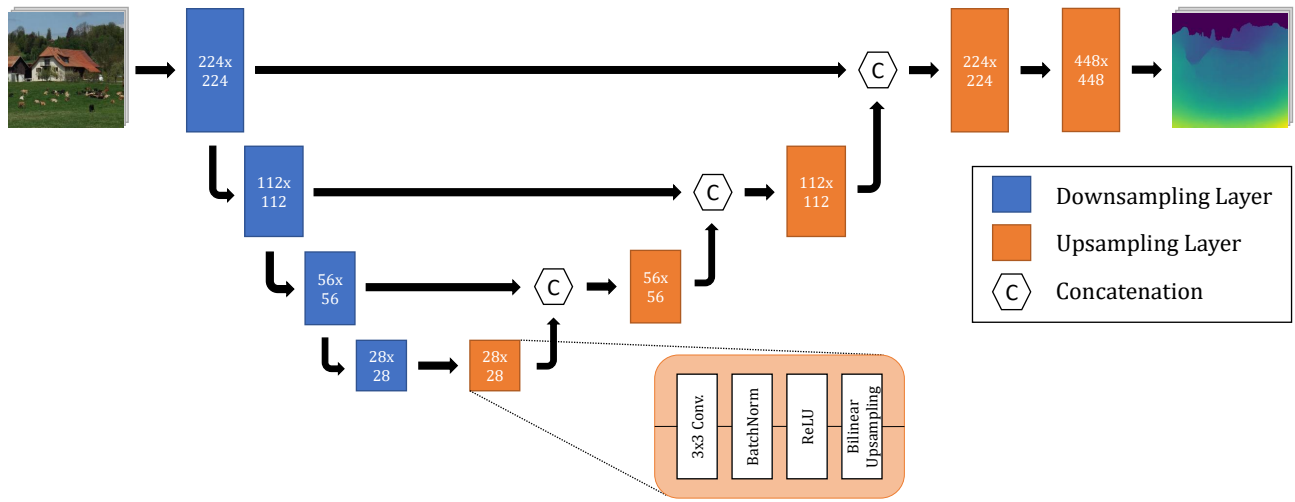

Figure 3. PLDepthEffNet U-net model architecture as proposed in the paper. The blue downsampling layers are specified by the used EfficientNet [11] backbone. The layer captions specify the corresponding output dimensionality of the respective layers.



\end{document}